\documentclass[a4paper,fleqn]{cas-dc}

\usepackage[authoryear,longnamesfirst]{natbib}

\usepackage[linesnumbered,ruled,vlined]{algorithm2e}
\usepackage{algpseudocode}
\usepackage{amsmath}

\def\tsc#1{\csdef{#1}{\textsc{\lowercase{#1}}\xspace}}
\tsc{WGM}
\tsc{QE}

\begin{document}
\let\WriteBookmarks\relax
\def\floatpagepagefraction{1}
\def\textpagefraction{.001}

\shorttitle{Attention-Based Lip Audio-Visual Synthesis for Talking Face Generation in the Wild}    

\shortauthors{Ganglai Wang et al.}  

\title [mode = title]{Attention-Based Lip Audio-Visual Synthesis for Talking Face Generation in the Wild}  

 

%

\author[a]{Ganglai Wang}


\fnmark[1]




\affiliation[a]{organization={Department of Computer Science},
            addressline={Northwestern Polytechnical University}, 
            city={Xi'an},
            country={China}}

\author[a]{Peng Zhang}
\cormark[1]
\fnmark[2]

\ead{zh0036ng@nwpu.edu.cn}



\affiliation[b]{organization={China Mobile-NCU AI\&IOT Jointed Lab, Informatization Office},
            addressline={Nanchang University}, 
            country={China}}
\author[a]{Lei Xie}
\fnmark[3]
\author[b]{Wei Huang}
\fnmark[4]
\author[a]{Yufei Zha}
\fnmark[5]




\begin{abstract}
Talking face generation with great practical significance has attracted more attention in recent audio-visual studies. How to achieve accurate lip synchronization is a long-standing challenge to be further investigated. Motivated by xxx, in this paper, an AttnWav2Lip model is proposed by incorporating spatial attention module and channel attention module into lip-syncing strategy. Rather than focusing on the unimportant regions of the face image, the proposed AttnWav2Lip model is able to pay more attention on the lip region reconstruction. To our limited knowledge, this is the first attempt to introduce attention mechanism to the scheme of talking face generation. An extensive experiments have been conducted to evaluate the effectiveness of the proposed model. Compared to the baseline measured by LSE-D and LSE-C metrics, a superior performance has been demonstrated on the benchmark lip synthesis datasets, including LRW, LRS2 and LRS3.
\end{abstract}




\maketitle

\section{Introduction}\label{Sec1}
Video, which is the most abundant information carrier, has become an indispensable part of information in our age. In order to present information more cordially to the audiences, the presentation of audio and video need to be matched or shown synchronously. Unfortunately, speech sound and lip movement are out of sync sometimes, which may put a damper on visual experience of viewers, even mislead them. Suppose there's a strong possibility that speech advances or lags behind the mouth due to device recording error, the failure of video later dubbing and video transmission may be inevitable. Beyond that, when videos are translated into other different languages, the original lip movements would mismatch the target speech very likely. Therefore, study of talking face generation, which aims to establish a corresponding relationship between the given input audio streams with the lip-sync talking face in the videos, has attracted great attention recently. Owing to its practical significance, this study can be used not only to correct the lip sync to match the current speech, but also to automatically create lips for animation characters to fully match the voice actors's speech, which is able to substantially save time and energy for animation producer.

Earlier introduced by \cite{16}, the study of lip synthesis has attracted increasingly interests of research in more recent vision based work. One important development is achieved by the multimodal fusion strategy with deep networks modeling. To learn a map from input mode to lips on a specified identity, \cite{2} proposed ObamaNet, in which they use the Char2Wav architecture (\cite{17}) to generate speech from the input text. Then, the speech synthesis system is trained using the audio and frames extracted from the videos. This approach can be employed to generate lips with specified identity for any text. \cite{3} firstly maps the audio features to sparse shape coefficients by RNNs, and then maps from shape to mouth texture/shapes, which finally synthesizes high detailed face textures. These works \cite{2,3} both require a large amount of data of a particular speaker for training, and it is also the reason why their model is only capable to synthesize specific identity and voice.

In different ways, \cite{1} learn a map from fusion of audio stream and video frames to target frames by using a simple fully convolutional encoder-decoder model. \cite{4} proposed to learn a joint audio-visual representation, then disentangle the subject-related and speech-related information using adversarial training to generate arbitrary subject talking faces. \cite{1,4} can achieve unconstrained talking face generation from any speech, but both of them are language-dependent, which means that they are limited to realize lip synchronization in specific language.

Training the model for different languages is not easy because collecting large video datasets in diverse languages is infeasible. From another point of view, the GAN-based \cite{32,33,34,35,36,37,38} approach should be promising for generating realistic talking faces conditioned on audio in any language such as LipGAN \cite{7}, which employs an adversary learning to measure the extent of lip synchronization in the frames generated by the generator. Besides, LipGAN inputs the target face with bottom-half masked in order to allow the generated mouth to be seamlessly pasted back into the original video without any post-processing.This mechanism works very well on static images but generate inaccurate lips while trying to lip-sync unconstrained videos in the wild.

To further overcome the limitation of LipGAN, \cite{15} proposed Wav2Lip model with two improvements: a buffer of T (T=5 in the original article) contiguous frames is utilized by the network to effectively make use of the temporal context information for lip-sync detection. Secondly, Wav2Lip employs a pre-trained lip-sync discriminator that is able to accurately detect sync in real videos to perform adversarial training of the generator. The reason for this is that lip region reconstruction loss corresponds to less than 4$\%$ of the total reconstruction loss (based on the spatial extent), while the discriminator of LipGAN mainly focuses on the visual artifacts instead of the audio-lip correspondence. Thus, to further improve the accuracy of lip synthesis, making the deep network pay more attention on the lip region is crucial, which motivates the Attention Mechanism \cite{18} being applied as the solution as it is in the vision and NLP tasks \cite{25,26,27,28,29,30}. The advantages of attention has been studied extensively in more recent literature \cite{9,10,11,20,21}. Attention mechanism tells the models where to focus, and enhances the representation of interests. \cite{19} proposed convolutional block attention module (CBAM) by ensembling spatial attention and channel attention, which can be seamlessly embedded in convolutional neural network. 

In this study, we incorporate an attention module into Wav2Lip to enable the network learn by itself `where to emphasize' and `where to suppress' in the feature maps across channel and spatial axes. The obtained lip-syncing model is named AttnWav2Lip, which has higher accuracy by embedding spatial attention and channel attention into Wav2Lip model. To verify the effectiveness and generalization of the proposed model, we only train the AttnWav2Lip model on the dataset LRS2, but evaluate it on LRS2 \cite{12}, LRS3 \cite{13}, LRW \cite{14} with the LSE-D and LSE-C \cite{15} metrics without any fine-tuning operation. Our main contributions contain three-fold:
\begin{itemize}
\item An attention-based lip-synchronization model `AttnWav2Lip' is proposed, which is able to achieve precise lip-syncing for arbitrary talking face videos in the wild with arbitrary speech.
\item The proposed strategy enables the lip-syncing network to pay more attention to lip region by embedding attention module, and result in substantially enhance the accuracy of lip-syncing from any given speech.
\item The effectiveness of the proposed method is validated using LSE-D and LSE-C metrics.
\end{itemize}

The rest of the paper is organized as follows: in section \ref{Sec2}, we briefly review the recent developments in talking face generation and introduce attention mechanism. Section \ref{Sec3} discusses the proposed work. Section \ref{Sec4} presents the experimental results with analysis. In section \ref{Sec5}, we study the contribution of different attention modules to the lip-syncing model.







\section{related work}\label{Sec2}
\noindent \textbf{Talking Face Generation. }The work of talking face generation from a given speech is a long-standing matter of great concern in multimedia applications. The related research work are usually divided into two categories: constrained speech-driven talking face generation and unconstrained speech-driven talking face generation. In constrained talking face synthesis \cite{2,3}, their lip-syncing model is either speaker-dependent (model training is on a small set of hours of  Barack Obama's presentation), or language-dependent (modeling training is on datasets with a limited set of words such as GRID \cite{23}, TIMIT \cite{24} and LRW \cite{14}). It is noteworthy that increasing related studies focus on designing a model  to lip-sync videos of arbitrary identities, voices, and languages. For instance, \cite{6} proposed Speech2Vid by using a joint embedding of the face and audio to generate synthesized talking face video frames based on an encoder-decoder convolutional neural network (CNN) model. The GAN-based LipGAN \cite{7} model inputs the target face with bottom-half masked to act as a pose prior, this guarantee that the generated face crops can seamlessly past back into the original video without further post-processing. As an extension of LipGAN, Wav2Lip \cite{15} employs a pre-trained lip-sync discriminator to correct the lip sync and utilizes a visual quality discriminator to improve the visual quality. Additionally, Wav2Lip processes a buffer of 5 contiguous frames at a time because a small temporal context is very helpful for detecting lip-sync, and the pre-trained lip-sync discriminator also needs a temporal window of same number of frames as input.\\

\noindent \textbf{Attention Mechanism. }While processing a picture, a normal way is to firstly scan the global image and then focus on the region that interests you. The human vision system \cite{40,41,42} makes you pay more attention on the details of the target and inhibit irrelevant information, which can greatly improve the efficiency and accuracy of visual information perception. Attention mechanism is a functionalility by imitating this human biological skill of visual observation, and has been widely used in different audio-visual applications. Typically, \cite{19} proposed convolutional block attention module(CBAM) to improve the representative ability of its network models. By sequentially inferring attention maps along channel and spatial axes, the attention maps of CBAM are then multiplied to the input feature map.\\

\noindent \textbf{Spatial Attention. }Convolutional neural network generates different dimensional feature maps from image by passing through convolution layers, polling layers and other modules. This process treats each position in the whole image equally. However, the difference between tasks requires the operation attention need to be focused on different regions, e.g. in image classification deep neural network pays more attention to target region and suppresses the background. In our work, emphasizing the lip region would make the network perform fine-grained lip shape correction as Wav2Lip did. Spatial attention mechanism generates a spatial attention map using the inter-spatial relationship of features, then the maps are multiplied to the input feature for telling where should be the informative part.\\

\noindent \textbf{Channel Attention. }Different channels from input images, convolutional blocks and residual skip connections represent different type of image information. Currently, most CNNs treat each channel in feature maps equally, which undoubtedly lack of discriminative learning ability across feature channels, and finally hinder the representational power of deep networks. In comparison to spatial attention, channel attention obtains a channel attention map by exploiting the inter-channel relationship of features that focuses on `what' is meaningful to model training. \\

\section{Attention-based lip audio-visual synthesis}\label{Sec3}
\begin{figure*}[htbp]
\centerline{\includegraphics [width=0.9\textwidth]{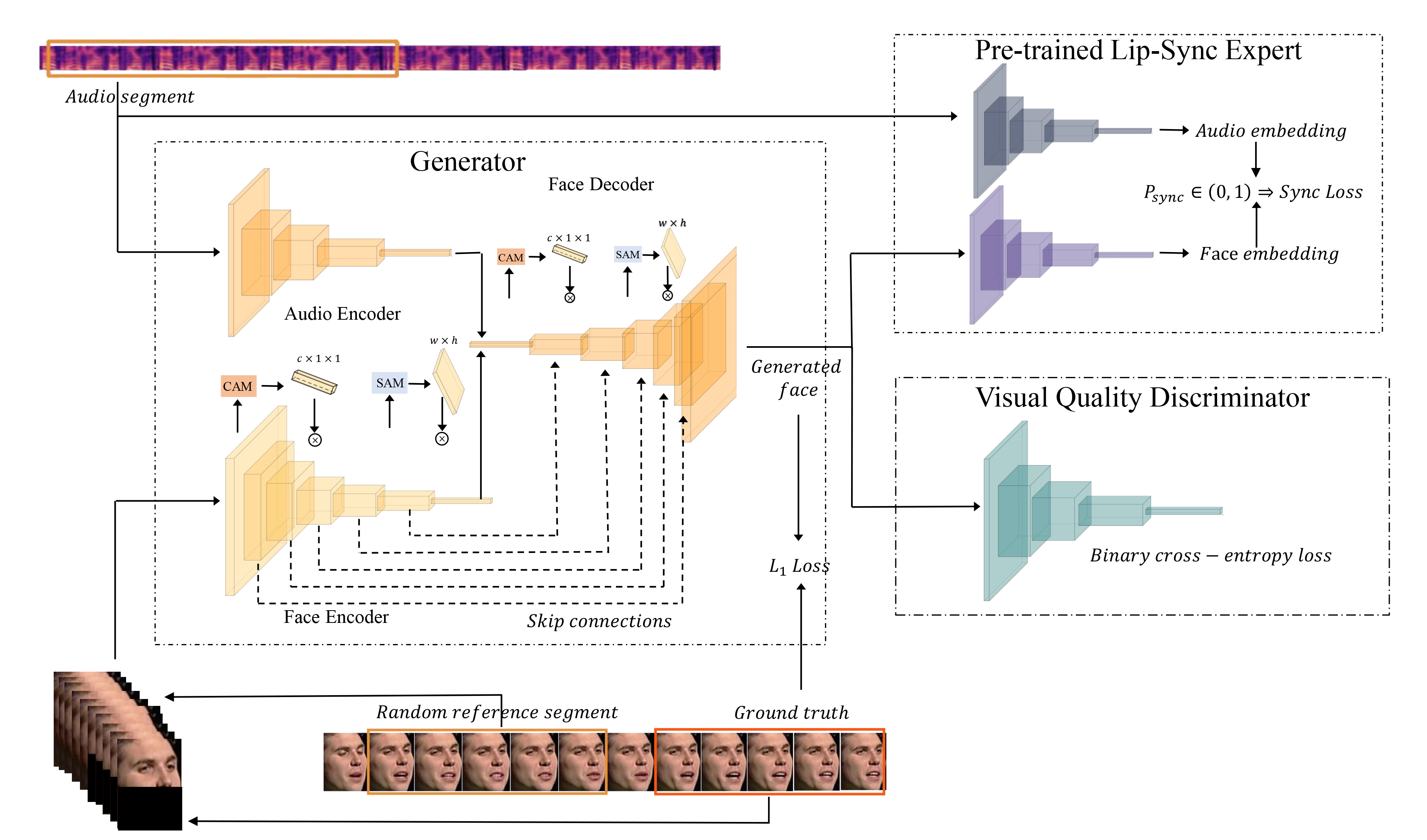}}
\caption{\textbf{Illustration of AttnWav2Lip. }Wav2Lip consists of a generator and two different discriminators (one for sync accuracy and another for better visual quality). The generator is composed of Audio Encoder, Face Encoder and Face Decoder, all of them are mainly developed by several blocks of convolution layers. The spatial attention module and channel attention module are sequentially embedded to the Face Encoder and the Face Decoder.}
\label{Net}
\end{figure*}

\subsection{Preliminary of Wav2Lip}
The model of Wav2Lip is composed by Generator, Visual Quality Discriminator, and Pre-trained Lip-Snyc Expert. The Generator $G$ contains three blocks: Face Encoder, Speech Encoder, and Face Decoder. The Face Encoder composed of a stack of residual convolutional layers encodes the input random reference frames, which are buffered (T=5) frames from a video concatenated with lower-half masked target-face along the channel axis. The Speech Encoder which encodes the input speech segment is a stack of convolutional layers as well. The Face Decoder that decodes the speech representation concatenated with the face embeddings into target-face frames is also a stack of convolutional layers, along with transpose convolutions for upsampling. For the generator, it is trained to minimize reconstruction loss $L_1$ between the generated frames $L_g$ and ground-truth frames $L_{gt}$ :

\begin{equation}
L_1 = \frac{1}{N} \sum_{i=1}^{N}||L_g - L_{gt}||_1
\label{1}
\end{equation}

\indent The principal innovative point of Wav2Lip is the well-trained lip-sync expert that is used to improve sync accuracy. This lip-sync discriminator inspired by Syncnet \cite{5} makes Generator spend more time to perform fine-grained lip shape correction instead of paying more attention on surrounding image reconstruction. Especially noted that the lip-sync expert processes $T=5$ contiguous frames at a time, and it is pre-trained whose parameters are not changed throughout the whole training process. The generator is also trained to minimize the expert sync-loss $L_{sync}$ from the expert discriminator:

\begin{equation}
\begin{split}
P_{sync} = \frac{E_v \cdot E_s}{max(||E_v||_2 \cdot ||E_s||_2)}\\
L_{sync} = \frac{1}{N} \sum_{i=1}^{N}-log(P_{sync}^i)
\end{split}
\label{sync}
\end{equation}
\indent where $P_{sync}$ is the probability denoting that the input audio-video pair is in sync. $E_v$ and $E_s$ correspond to video and speech embeddings generated by the pre-trained lip-snyc expert.

\indent Using a strong lip-sync discriminator forces the generator to produce accurate lip shapes. However, it sometimes results in the deformed regions to be slightly blurry or contain slight artifacts. To solve this, Wav2Lip trains a visual quality discriminator in a GAN setup along with the generator. The visual quality discriminator D whose input is only the lower half of the generated face which consists of a stack of convolutional blocks. The training of D is to maximize the objective function $L_{disc}$(Equation \ref{disc}):

\begin{equation}
\begin{split}
L_{gen} = E_{x~L_g}[log(1 - D(x))]\\
L_{disc} = E_{x~L_{gt}}[log(D(x))] + L_{gen}
\end{split}
\label{disc}
\end{equation}

\indent where $L_g$ corresponds to the images from the generator G, and $L_{gt}$ corresponds to the real images.

For generator, The minimization of Equation \ref{L} is the weighted sum of the reconstruction loss $L_1$(Equation \ref{1}), the synchronization loss $L_{sync}$(Equation \ref{sync}) and the adversarial loss $L_{gen}$(Equation \ref{disc}):

\begin{equation}
L = (1 - \alpha - \beta) \cdot L_{1} + \alpha \cdot L_{sync} + \beta \cdot L_{gen}
\label{L}
\end{equation}

\indent where $\alpha$ is the synchronization penalty weight $\beta$ is the adversarial loss, which are empirically set as $\alpha=0.03$ and $\beta=0.07$ in Wav2Lip. 

\indent No matter does Wav2Lip use a pre-trained lip-sync expert for sync accuracy, or input the lower half of the generated face to the visual quality discriminator, the same target is to make the network perform attentively fine-grained lip shape correction. One effective way is the attention mechanism, which enables the networks focus on important features and suppress unnecessary ones. Next, we discuss in detail how the proposed work can achieve this purpose.

\begin{figure}[htbp]
\centerline{\includegraphics [width=0.5\textwidth]{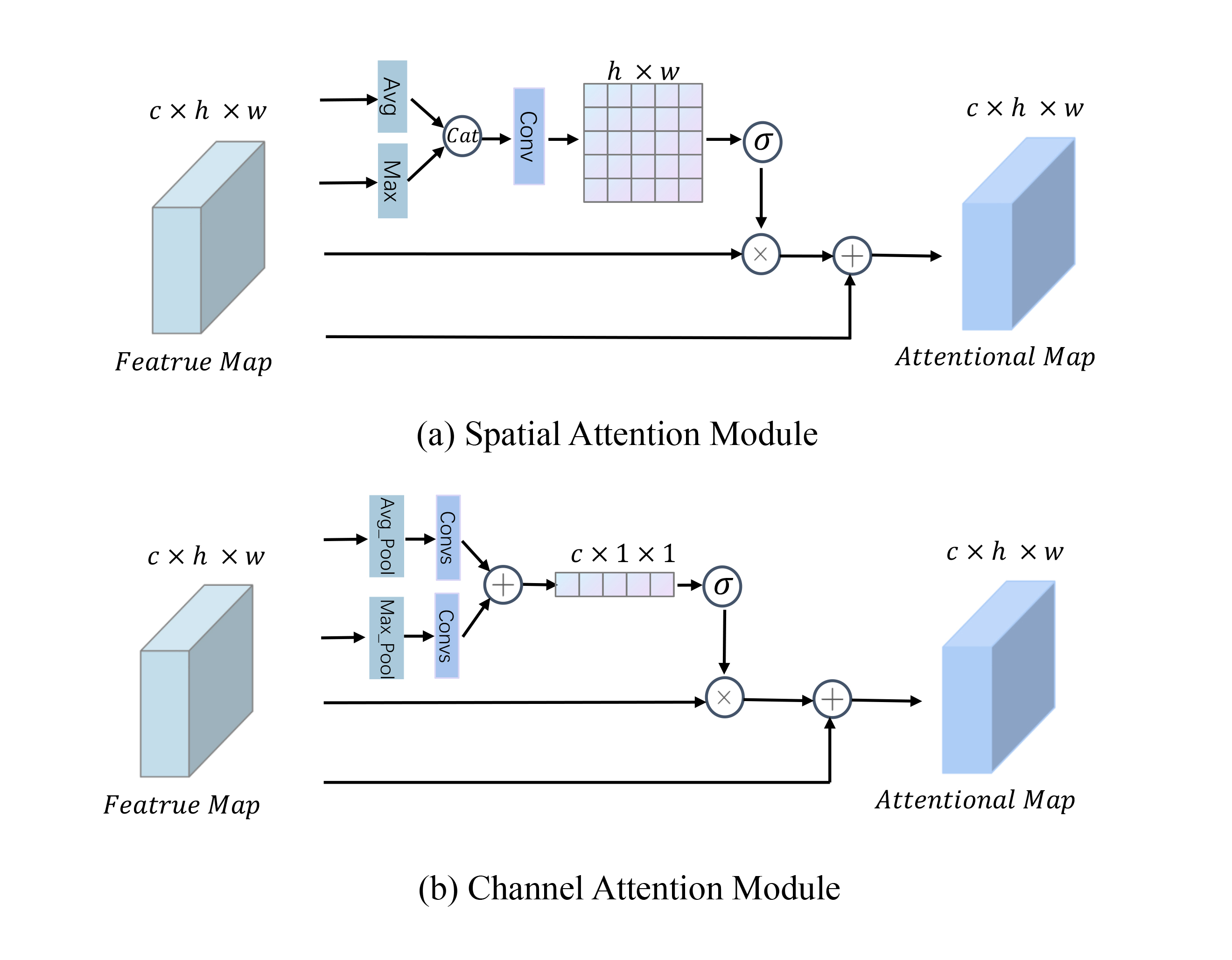}}
\caption{ \textbf{Illustration of attention module. }The spatial attention module is illustrated on the top, and channel attention module on the bottom.}
\label{Attn}
\end{figure}

\subsection{Attention Based Wav2Lip}
Attention module $Attn$ infers an attention map $W$ for the given feature map $F$, then the attentional map $F^{\prime}$ is obtained by multiplying feature map and attention map using Equation \ref{F1}. To introduce this work,  we firstly elaborate the spatial attention and channel attention respectively, and then discuss how to use them in Wav2Lip.

\begin{equation}
F^{\prime}= F \,\otimes\, Attn(F)
\label{F1}
\end{equation}
\indent where $\otimes$ denotes element-wise multiplication.\\

\noindent \textbf{Spatial Attention Module (SAM). }Spatial attention module infers a spatial attention map which decides where to emphasize or suppress. To compute the spatial attention map, the operations of average-pooling and max-pooling need to be firstly performed along the channel axis on the feature map $F$, and the obtained results are then concatenated to generate an intervening feature. The pooling operations along the channel axis has been verified effectively in highlighting informative regions \cite{19}. Following concatenation, a convolution layer and a sigmoid layer are applied successively to generate a spatial attention map $M_s$ as Equation \ref{M_s}. Fig. \ref{Attn}(a) illustrates this process of spatial attention map.

\begin{equation}
M_s= \sigma (Conv(Concat(AvgPool(F), AvgPool(F))))
\label{M_s}
\end{equation}

\indent where $AvgPool$ and $AvgPool$ correspond to average-pooling and max-pooling operations, $concat$ is concatenation operation, $Conv$ represents a convolution operation with the filter size of $7 \times 7$ or $3 \times 3$, and $\sigma$ denotes the sigmoid function.\\

\noindent \textbf{Channel Attention Module (CAM). }A channel attention map is generated by exploiting the inter-channel relationship of features, which tells `what' is meaningful to the input feature. An intuitive interpretation is that channel attention module infers weight for each channel in the feature map. We first use average-pooling and max-pooling operations to generate two different spatial context: average-pooled features and max-pooled features \cite{19} and then both of them are forwarded to a shared convolution layers followed by a sigmoid layer to produce channel attention map $M_c$ as Equation \ref{M_c}. Fig. \ref{Attn}(b) depicts the computation process of channel attention map.

\begin{equation}
M_c= \sigma (Convs(Avg\_Pool(F) \,\oplus\, Convs(Avg\_Pool(F))))
\label{M_c}
\end{equation}

\indent where $Avg\_Pool$ and $Avg\_Pool$ denote average-pooling and max-pooling respectively, $\oplus$ denotes the element-wise addition, $Convs$ represents convolution operations, and $\sigma$ denotes the sigmoid function.\\

\noindent \textbf{Attention Based Wav2Lip. }We have obtained spatial attention map $M_s$ and channel attention map $M_c$ focusing on `where' and `what' respectively. The arrangement of the channel attention followed by the spatial attention performs best, which has been demonstrated by experiments \cite{19}. Fig. \ref{Net} shows the network architecture of the proposed AttnWav2Lip. We embed SAM and CAM to each block of the Face Encoder (7 blocks) and Face Decode (7 blocks). Besides, the residual structure is also applied to attention blocks. However, we do not directly apply attention modules to the input images not processed by convolution layers, because the pixels of input map with high resolution have extremely weak semantic information, which is conflict with the region properties of attention mechanism. For a given feature map $F$, the attentional map can be generated $F^{\prime\prime}$ by Equation \ref{F2}. The detailed training of of AttnWav2Lip model is illustrated in  Algorithm 1.

\begin{equation}
\begin{split}
F^\prime = F \,\oplus\, F \,\otimes\, CAM(F)\\
F^{\prime\prime} = F^\prime \,\oplus\, F^\prime \,\otimes\, SAM(F^\prime)
\end{split}
\label{F2}
\end{equation}
\indent where $\otimes$ , $\oplus$ denote element-wise multiplication and addtion respectively, $F^\prime$ represents the channel attentional map.

\begin{table*}[t]
\caption{The quantitative results on LRW \cite{14}, LRS2 \cite{12}, LRS3 \cite{13}}
$$
\setlength{\arraycolsep}{1mm}{
\begin{array}{lcccccccc}
\toprule[1.5pt]
\rule{0pt}{6pt}
& \multicolumn{2}{c} {\text {    LRW   }} & & \multicolumn{2}{c} {\text {   LRS2   }} & & \multicolumn{2}{c} {\text {   LRS3   }} \\
\cline { 2 - 3 } \cline { 5 - 6 }  \cline{ 8 - 9}
\rule{0pt}{10pt}
\text { Method } & \text {LSE-D}\downarrow & \text {LSE-C}\uparrow & & \text {LSE-D}\downarrow & \text {LSE-C}\uparrow & & \text {LSE-D}\downarrow & \text {LSE-C}\uparrow \\
\hline
\rule{0pt}{10pt}
\textbf {Speech2Vid \cite{6} } & 13.14 & 1.762 & & 14.23 & 1.587 & & 13.97 & 1.681\\
\rule{0pt}{10pt}
\textbf {LipGAN \cite{7} } & 10.05 & 3.350 & & 10.33 & 3.199 & & 10.65 & 3.193\\
\rule{0pt}{10pt}
\textbf {Wav2Lip \cite{15} } & 8.268 & \textbf{5.758} & & 7.521 & 6.406 & & \textbf{8.835} & \textbf{5.446}\\
\rule{0pt}{10pt}
\textbf {AttnWav2Lip(ours)} & \textbf{8.245} & 5.663 & & \textbf{7.339} & \textbf{6.530} & & 8.934 & 5.226\\
\bottomrule[1.5pt]
\end{array}
}
$$
\label{Table1}
\end{table*}

\begin{algorithm}
  \caption{Traing process of AttnWav2Lip}
  \KwIn{\textbf{F}, five random face frames from the video concatenated with target-face with lower-half masked along the channel axis. \textbf{S}, the input speech segment corresponding to target-face frames.}
  \KwOut{\textbf{G}, five face frames matching with input speech.}
  \For{$i \le epochs$}
  {
    $S^\prime = Speech Edcoder(S) $\;
    $List = []$\;
    \While{$Block$ in $Face Encoder$}
    {
      $F = Block(F)$\;
      $F = F + F \otimes CAM(F)$\;
      $F = F + F \otimes SAM(F)$\;
      $F \longrightarrow List$\;
     }
    \While{$Block$ in $Face Dncoder$}
    {
      $F^\prime = Concatenate(S^\prime, List[-1])$\;
      $F^\prime = Block(F^\prime)$\;
      $F^\prime = F^\prime + F^\prime \otimes CAM(F^\prime)$\;
      $F^\prime = F^\prime + F^\prime \otimes SAM(F^\prime)$\;
      $List.pop()$\;
     }
     Get loss $L_{disc}$ according to equation(\ref{disc}).\;
     Get loss $L$ according to equation(\ref{L}).\;
     Minimize $L$ to update $Generator$.\;
     Minimize $L_{disc}$ to update $Visual\,Quality\,Discriminator$.\;
  }
\label{Algorithm1}
\end{algorithm}

\section{Experiments}\label{Sec4}

\noindent \textbf{Dataset. }The proposed AttnWa2Lip model is only trained on the LRS2 dataset \cite{12} and evaluated on the LRS2, LRS3 \cite{13}, and LRW \cite{14}. The LRS2 dataset contains over 29 hours of talking faces from BBC TV program, we use more than 45000 videos for training and 1120 for evaluation. The LRW dataset is composed of up to 1000 utterances, spoken by hundreds of different speakers from TV broadcasts, we use 2000 videos for evaluation. The LRS3 dataset consists of thousands of spoken sentences from TED and TEDx videos, and 1320 videos are used for evaluation. Noted that for each testing video, a audio is taken from another randomly-sampled video with the condition that the length of the speech is less than the video.\\

\noindent \textbf{Metrics. }The Metrics of LSE-D (Lip Sync Error - Distance) and LSE-C (Lip Sync Error - Confidence) for evaluating lip-sync in the wild were proposed in \cite{15}. LSE-D is calculated in terms of the distance between the lip and audio representations both generated by the pre-trained SyncNet \cite{5}, and the lower the LSE-D, the higher the audio-visual matching. LSE-C refers to the average confidence score, the higher the LSE-C, the better the audio-video correlation. The effectiveness of proposed AttnWav2Lip is validated on both of LSE-D and LSE-C metrics.\\

\noindent \textbf{Implementation Details. }The AttnWav2Lip is implemented using Pytorch. The initial face detection is performed using S3FD \cite{22} and the obtained face crops are resized to $96 \times 96 \times 3$. We use the same experiment setup as Wav2Lip with a batch size of 80, a learning rate of $1e^{-4}$ and using the Adam \cite{39} optimizer. It should be noted that in Wav2Lip, the synchronization penalty weight $\alpha$ is set to zero in the beginning, then $\alpha$ will be automatically set to 0.03 when the model evaluation lip-sync loss $L_{sync}$ is down to 0.75. However, when training the Wav2Lip model in this work, it is found find that $L_{sync}$ is almost impossible to reduce to 0.75, which forces the the training to be stopped when the reconstruction loss $L_1$ is no longer dropping. By manually set the synchronization penalty weight $\alpha$ to 0.03, the model is then fine-tuned from the last checkpoints. It takes less then 2 days to finalize the whole training process. \\

\noindent \textbf{Qualitative Results. }The proposed work is compared with three typical models of unconstrained talking face generation from given speech, which are: Speech2Vid \cite{6}, an encoder-decoder convolutional neural network (CNN) model; LipGAN \cite{7}, a reconstruction-based method; Wav2Lip \cite{15},  an extended version of LipGAN and as our baseline. An adequate comparison is shown in Table \ref{Table1}. Note that we only train on the train set on LRS2 \cite{12}, but generate lips on other datasets LRW \cite{14} and LRS3 \cite{13} without any further fine-tuning. The comparison results show that the proposed work outperforms the other three approaches, which demonstrates the effectiveness of embedding attention module to the model of talking face generation from speech.

\section{Ablation studies}\label{Sec5}

\begin{table}[b]
\caption{\textbf{Ablation study with quantitative comparisons on AttnWav2Lip. } }
$$
\begin{array}{lcc}
\toprule[1.5pt]
\rule{0pt}{6pt}
\text { Method } & \text { LSE-D } \downarrow & \text { LSE-C } \uparrow\\
\hline
\rule{0pt}{10pt}
\text {Wav2Lip \cite{15} } & 7.521 & 6.406\\
\rule{0pt}{10pt}
\text { SattnWav2Lip } & 7.358 & 6.415\\
\rule{0pt}{10pt}
\text { CattnWav2Lip } & 7.430 & 6.405\\
\rule{0pt}{10pt}
\textbf { AttnWav2Lip } & \textbf{7.339} & \textbf{6.530}\\
\bottomrule[1.5pt]
\end{array}
$$
\label{Table2}
\end{table}

\begin{figure}[htbp]
\centerline{\includegraphics [width=0.4\textwidth]{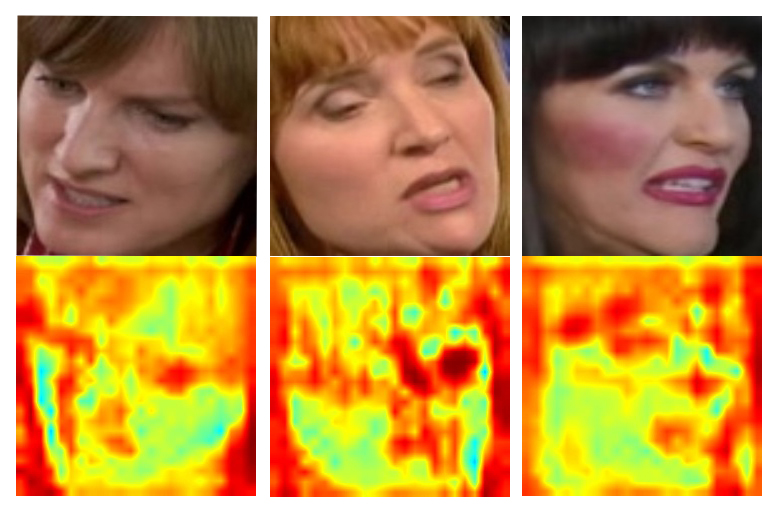}}
\caption{ \textbf{Visualization of spatial attention map.}The brighter the color, the higher the weight. It shows the network focus on different regions inequality, apparently, pay more attention on lip region.}
\label{Visual}
\end{figure}

The ablation studies are conducted on two attention modules in the proposed work. To study the contributions of SAM and CAM to Wav2Lip, the experiments on our model contain: (1)SattnWav2Lip: only embedding spatial attention module with residual connection to convolution blocks; (2)CattnWav2Lip: only embedding channel attention module with residual connection to convolution blocks; (3)SattnWav2Lip. As shown in Table \ref{Table2}, either of the spatial attention module or the channel attention module can improve lip-sync accuracy measured by LSE-D and LSE-C metrics. However, the contribution of spatial attention is more than that of channel attention. As stated in section \ref{Sec1}, that the model pays more attention to the lip region is the key to enhance the accuracy of lip-syncing, and this is just what the spatial attention does. Fig. \ref{Visual} presents the visualization of some spatial attention maps, in which the brighter the color, the higher the weight. Inside a face image, it is clearly shown that the weight of lip region is larger than other regions, this means that these areas have been emphasized during training to enable the generator pay more attention on lip shape reconstruction.

\section{Conclusion}\label{6}
In this paper, we proposed an attention based lip synthesis model for talking face generation called AttnWav2Lip. The proposed model is designed to learn "where to emphasize" and "what is more important" by embedding spatial attention module and channel attention module into convolution blocks of Wav2Lip. To the best of our knowledge, this is the first time that attention mechanism is utilized in the work of lip-syncing from a given speech. Based on a substantial experiments using the lip-syncing metrics of LSE-D and LSE-C, the effectiveness of the proposed work has been convincibly validated.

\section*{Acknowledgments}
This work was jointly supported by grants 61971352 \& 61862043 \& 61773397 \& 61801392 approved by National Natural Science Foundation of China. The key grant 20204BCJ22011 approved by Natural Science Foundation of Jiangxi Province in China, the equipment pre research project 61400010115, as well as the National Science Foundation for Young Scientists of China 61801392.

\bibliographystyle{cas-model2-names}

\bibliography{mybibfile}

\begin{thebibliography}{40}
\expandafter\ifx\csname natexlab\endcsname\relax\def\natexlab#1{#1}\fi
\providecommand{\url}[1]{\texttt{#1}}
\providecommand{\href}[2]{#2}
\providecommand{\path}[1]{#1}
\providecommand{\DOIprefix}{doi:}
\providecommand{\ArXivprefix}{arXiv:}
\providecommand{\URLprefix}{URL: }
\providecommand{\Pubmedprefix}{pmid:}
\providecommand{\doi}[1]{\href{http://dx.doi.org/#1}{\path{#1}}}
\providecommand{\Pubmed}[1]{\href{pmid:#1}{\path{#1}}}
\providecommand{\bibinfo}[2]{#2}
\ifx\xfnm\relax \def\xfnm[#1]{\unskip,\space#1}\fi
\bibitem[{Afouras et~al.(2018a)Afouras, Chung, Senior, Vinyals and
  Zisserman}]{12}
\bibinfo{author}{Afouras, T.}, \bibinfo{author}{Chung, J.S.},
  \bibinfo{author}{Senior, A.}, \bibinfo{author}{Vinyals, O.},
  \bibinfo{author}{Zisserman, A.}, \bibinfo{year}{2018}a.
\newblock \bibinfo{title}{Deep audio-visual speech recognition}.
\newblock \bibinfo{journal}{IEEE transactions on pattern analysis and machine
  intelligence} .
\bibitem[{Afouras et~al.(2018b)Afouras, Chung and Zisserman}]{13}
\bibinfo{author}{Afouras, T.}, \bibinfo{author}{Chung, J.S.},
  \bibinfo{author}{Zisserman, A.}, \bibinfo{year}{2018}b.
\newblock \bibinfo{title}{Lrs3-ted: a large-scale dataset for visual speech
  recognition}.
\newblock \bibinfo{journal}{ArXiv} \bibinfo{volume}{abs/1809.00496}.
\bibitem[{Ba et~al.(2015)Ba, Mnih and Kavukcuoglu}]{10}
\bibinfo{author}{Ba, J.}, \bibinfo{author}{Mnih, V.},
  \bibinfo{author}{Kavukcuoglu, K.}, \bibinfo{year}{2015}.
\newblock \bibinfo{title}{Multiple object recognition with visual attention}.
\newblock \bibinfo{journal}{CoRR} \bibinfo{volume}{abs/1412.7755}.
\bibitem[{Bahdanau et~al.(2015a)Bahdanau, Cho and Bengio}]{25}
\bibinfo{author}{Bahdanau, D.}, \bibinfo{author}{Cho, K.},
  \bibinfo{author}{Bengio, Y.}, \bibinfo{year}{2015}a.
\newblock \bibinfo{title}{Neural machine translation by jointly learning to
  align and translate}.
\newblock \bibinfo{journal}{CoRR} \bibinfo{volume}{abs/1409.0473}.
\bibitem[{Bahdanau et~al.(2015b)Bahdanau, Cho and Bengio}]{21}
\bibinfo{author}{Bahdanau, D.}, \bibinfo{author}{Cho, K.},
  \bibinfo{author}{Bengio, Y.}, \bibinfo{year}{2015}b.
\newblock \bibinfo{title}{Neural machine translation by jointly learning to
  align and translate}.
\newblock \bibinfo{journal}{CoRR} \bibinfo{volume}{abs/1409.0473}.
\bibitem[{Bregler et~al.(1997)Bregler, Covell and Slaney}]{16}
\bibinfo{author}{Bregler, C.}, \bibinfo{author}{Covell, M.},
  \bibinfo{author}{Slaney, M.}, \bibinfo{year}{1997}.
\newblock \bibinfo{title}{Video rewrite: driving visual speech with audio}.
\newblock \bibinfo{journal}{Proceedings of the 24th annual conference on
  Computer graphics and interactive techniques} .
\bibitem[{Chung et~al.(2017)Chung, Jamaludin and Zisserman}]{1}
\bibinfo{author}{Chung, J.S.}, \bibinfo{author}{Jamaludin, A.},
  \bibinfo{author}{Zisserman, A.}, \bibinfo{year}{2017}.
\newblock \bibinfo{title}{You said that?}
\newblock \bibinfo{journal}{ArXiv} \bibinfo{volume}{abs/1705.02966}.
\bibitem[{Chung and Zisserman(2016a)}]{14}
\bibinfo{author}{Chung, J.S.}, \bibinfo{author}{Zisserman, A.},
  \bibinfo{year}{2016}a.
\newblock \bibinfo{title}{Lip reading in the wild}, in:
  \bibinfo{booktitle}{ACCV}.
\bibitem[{Chung and Zisserman(2016b)}]{5}
\bibinfo{author}{Chung, J.S.}, \bibinfo{author}{Zisserman, A.},
  \bibinfo{year}{2016}b.
\newblock \bibinfo{title}{Out of time: Automated lip sync in the wild}, in:
  \bibinfo{booktitle}{ACCV Workshops}.
\bibitem[{Cooke et~al.(2006)Cooke, Barker, Cunningham and Shao}]{23}
\bibinfo{author}{Cooke, M.}, \bibinfo{author}{Barker, J.},
  \bibinfo{author}{Cunningham, S.}, \bibinfo{author}{Shao, X.},
  \bibinfo{year}{2006}.
\newblock \bibinfo{title}{An audio-visual corpus for speech perception and
  automatic speech recognition.}
\newblock \bibinfo{journal}{The Journal of the Acoustical Society of America}
  \bibinfo{volume}{120 5 Pt 1}, \bibinfo{pages}{2421--4}.
\bibitem[{Corbetta and Shulman(2002)}]{42}
\bibinfo{author}{Corbetta, M.}, \bibinfo{author}{Shulman, G.},
  \bibinfo{year}{2002}.
\newblock \bibinfo{title}{Control of goal-directed and stimulus-driven
  attention in the brain}.
\newblock \bibinfo{journal}{Nature Reviews Neuroscience} \bibinfo{volume}{3},
  \bibinfo{pages}{201--215}.
\bibitem[{Goodfellow et~al.(2014)Goodfellow, Pouget-Abadie, Mirza, Xu,
  Warde-Farley, Ozair, Courville and Bengio}]{32}
\bibinfo{author}{Goodfellow, I.J.}, \bibinfo{author}{Pouget-Abadie, J.},
  \bibinfo{author}{Mirza, M.}, \bibinfo{author}{Xu, B.},
  \bibinfo{author}{Warde-Farley, D.}, \bibinfo{author}{Ozair, S.},
  \bibinfo{author}{Courville, A.}, \bibinfo{author}{Bengio, Y.},
  \bibinfo{year}{2014}.
\newblock \bibinfo{title}{Generative adversarial networks}.
\newblock \href{http://arxiv.org/abs/1406.2661}{\tt arXiv:1406.2661}.
\bibitem[{Harte and Gillen(2015)}]{24}
\bibinfo{author}{Harte, N.}, \bibinfo{author}{Gillen, E.},
  \bibinfo{year}{2015}.
\newblock \bibinfo{title}{Tcd-timit: An audio-visual corpus of continuous
  speech}.
\newblock \bibinfo{journal}{IEEE Transactions on Multimedia}
  \bibinfo{volume}{17}, \bibinfo{pages}{603--615}.
\bibitem[{Isola et~al.(2017)Isola, Zhu, Zhou and Efros}]{35}
\bibinfo{author}{Isola, P.}, \bibinfo{author}{Zhu, J.Y.},
  \bibinfo{author}{Zhou, T.}, \bibinfo{author}{Efros, A.A.},
  \bibinfo{year}{2017}.
\newblock \bibinfo{title}{Image-to-image translation with conditional
  adversarial networks}.
\newblock \bibinfo{journal}{2017 IEEE Conference on Computer Vision and Pattern
  Recognition (CVPR)} , \bibinfo{pages}{5967--5976}.
\bibitem[{Itti et~al.(2009)Itti, Koch and Niebur}]{40}
\bibinfo{author}{Itti, L.}, \bibinfo{author}{Koch, C.},
  \bibinfo{author}{Niebur, E.}, \bibinfo{year}{2009}.
\newblock \bibinfo{title}{A model of saliency-based visual attention for rapid
  scene analysis}.
\newblock \bibinfo{journal}{IEEE Trans. Pattern Anal. Mach. Intell.}
  \bibinfo{volume}{20}, \bibinfo{pages}{1254--1259}.
\bibitem[{Jaderberg et~al.(2015)Jaderberg, Simonyan, Zisserman and
  Kavukcuoglu}]{20}
\bibinfo{author}{Jaderberg, M.}, \bibinfo{author}{Simonyan, K.},
  \bibinfo{author}{Zisserman, A.}, \bibinfo{author}{Kavukcuoglu, K.},
  \bibinfo{year}{2015}.
\newblock \bibinfo{title}{Spatial transformer networks}, in:
  \bibinfo{booktitle}{NIPS}.
\bibitem[{Jamaludin et~al.(2019)Jamaludin, Chung and Zisserman}]{6}
\bibinfo{author}{Jamaludin, A.}, \bibinfo{author}{Chung, J.S.},
  \bibinfo{author}{Zisserman, A.}, \bibinfo{year}{2019}.
\newblock \bibinfo{title}{You said that?: Synthesising talking faces from
  audio}.
\newblock \bibinfo{journal}{International Journal of Computer Vision} ,
  \bibinfo{pages}{1--13}.
\bibitem[{K~R et~al.(2019)K~R, Mukhopadhyay, Philip, Jha, Namboodiri and
  Jawahar}]{7}
\bibinfo{author}{K~R, P.}, \bibinfo{author}{Mukhopadhyay, R.},
  \bibinfo{author}{Philip, J.}, \bibinfo{author}{Jha, A.},
  \bibinfo{author}{Namboodiri, V.}, \bibinfo{author}{Jawahar, C.V.},
  \bibinfo{year}{2019}.
\newblock \bibinfo{title}{Towards automatic face-to-face translation}.
\newblock \bibinfo{journal}{Proceedings of the 27th ACM International
  Conference on Multimedia} \URLprefix
  \url{http://dx.doi.org/10.1145/3343031.3351066},
  \DOIprefix\doi{10.1145/3343031.3351066}.
\bibitem[{Karras et~al.(2018)Karras, Aila, Laine and Lehtinen}]{37}
\bibinfo{author}{Karras, T.}, \bibinfo{author}{Aila, T.},
  \bibinfo{author}{Laine, S.}, \bibinfo{author}{Lehtinen, J.},
  \bibinfo{year}{2018}.
\newblock \bibinfo{title}{Progressive growing of gans for improved quality,
  stability, and variation}.
\newblock \bibinfo{journal}{ArXiv} \bibinfo{volume}{abs/1710.10196}.
\bibitem[{Kingma and Ba(2015)}]{39}
\bibinfo{author}{Kingma, D.P.}, \bibinfo{author}{Ba, J.}, \bibinfo{year}{2015}.
\newblock \bibinfo{title}{Adam: A method for stochastic optimization}.
\newblock \bibinfo{journal}{CoRR} \bibinfo{volume}{abs/1412.6980}.
\bibitem[{Kumar et~al.(2018)Kumar, Sotelo, Kumar, Br{\'e}bisson and Bengio}]{2}
\bibinfo{author}{Kumar, R.}, \bibinfo{author}{Sotelo, J.M.R.},
  \bibinfo{author}{Kumar, K.}, \bibinfo{author}{Br{\'e}bisson, A.D.},
  \bibinfo{author}{Bengio, Y.}, \bibinfo{year}{2018}.
\newblock \bibinfo{title}{Obamanet: Photo-realistic lip-sync from text}.
\newblock \bibinfo{journal}{ArXiv} \bibinfo{volume}{abs/1801.01442}.
\bibitem[{Luong et~al.(2015)Luong, Pham and Manning}]{26}
\bibinfo{author}{Luong, T.}, \bibinfo{author}{Pham, H.},
  \bibinfo{author}{Manning, C.D.}, \bibinfo{year}{2015}.
\newblock \bibinfo{title}{Effective approaches to attention-based neural
  machine translation}.
\newblock \bibinfo{journal}{ArXiv} \bibinfo{volume}{abs/1508.04025}.
\bibitem[{Mirza and Osindero(2014)}]{33}
\bibinfo{author}{Mirza, M.}, \bibinfo{author}{Osindero, S.},
  \bibinfo{year}{2014}.
\newblock \bibinfo{title}{Conditional generative adversarial nets}.
\newblock \bibinfo{journal}{ArXiv} \bibinfo{volume}{abs/1411.1784}.
\bibitem[{Mnih et~al.(2014)Mnih, Heess, Graves and Kavukcuoglu}]{9}
\bibinfo{author}{Mnih, V.}, \bibinfo{author}{Heess, N.},
  \bibinfo{author}{Graves, A.}, \bibinfo{author}{Kavukcuoglu, K.},
  \bibinfo{year}{2014}.
\newblock \bibinfo{title}{Recurrent models of visual attention}, in:
  \bibinfo{booktitle}{NIPS}.
\bibitem[{PrajwalK et~al.(2020)PrajwalK, Mukhopadhyay, Namboodiri and
  Jawahar}]{15}
\bibinfo{author}{PrajwalK, R.}, \bibinfo{author}{Mukhopadhyay, R.},
  \bibinfo{author}{Namboodiri, V.}, \bibinfo{author}{Jawahar, C.V.},
  \bibinfo{year}{2020}.
\newblock \bibinfo{title}{A lip sync expert is all you need for speech to lip
  generation in the wild}.
\newblock \bibinfo{journal}{Proceedings of the 28th ACM International
  Conference on Multimedia} .
\bibitem[{Radford et~al.(2016)Radford, Metz and Chintala}]{34}
\bibinfo{author}{Radford, A.}, \bibinfo{author}{Metz, L.},
  \bibinfo{author}{Chintala, S.}, \bibinfo{year}{2016}.
\newblock \bibinfo{title}{Unsupervised representation learning with deep
  convolutional generative adversarial networks}.
\newblock \bibinfo{journal}{CoRR} \bibinfo{volume}{abs/1511.06434}.
\bibitem[{Rensink(2000)}]{41}
\bibinfo{author}{Rensink, R.A.}, \bibinfo{year}{2000}.
\newblock \bibinfo{title}{The dynamic representation of scenes}.
\newblock \bibinfo{journal}{Visual Cognition} \bibinfo{volume}{7},
  \bibinfo{pages}{17 -- 42}.
\bibitem[{Sotelo et~al.(2017)Sotelo, Mehri, Kumar, Santos, Kastner, Courville
  and Bengio}]{17}
\bibinfo{author}{Sotelo, J.M.R.}, \bibinfo{author}{Mehri, S.},
  \bibinfo{author}{Kumar, K.}, \bibinfo{author}{Santos, J.F.},
  \bibinfo{author}{Kastner, K.}, \bibinfo{author}{Courville, A.C.},
  \bibinfo{author}{Bengio, Y.}, \bibinfo{year}{2017}.
\newblock \bibinfo{title}{Char2wav: End-to-end speech synthesis}, in:
  \bibinfo{booktitle}{ICLR}.
\bibitem[{Suwajanakorn et~al.(2017)Suwajanakorn, Seitz and
  Kemelmacher-Shlizerman}]{3}
\bibinfo{author}{Suwajanakorn, S.}, \bibinfo{author}{Seitz, S.},
  \bibinfo{author}{Kemelmacher-Shlizerman, I.}, \bibinfo{year}{2017}.
\newblock \bibinfo{title}{Synthesizing obama}.
\newblock \bibinfo{journal}{ACM Transactions on Graphics (TOG)}
  \bibinfo{volume}{36}, \bibinfo{pages}{1 -- 13}.
\bibitem[{Vaswani et~al.(2017a)Vaswani, Shazeer, Parmar, Uszkoreit, Jones,
  Gomez, Kaiser and Polosukhin}]{18}
\bibinfo{author}{Vaswani, A.}, \bibinfo{author}{Shazeer, N.M.},
  \bibinfo{author}{Parmar, N.}, \bibinfo{author}{Uszkoreit, J.},
  \bibinfo{author}{Jones, L.}, \bibinfo{author}{Gomez, A.N.},
  \bibinfo{author}{Kaiser, L.}, \bibinfo{author}{Polosukhin, I.},
  \bibinfo{year}{2017}a.
\newblock \bibinfo{title}{Attention is all you need}.
\newblock \bibinfo{journal}{ArXiv} \bibinfo{volume}{abs/1706.03762}.
\bibitem[{Vaswani et~al.(2017b)Vaswani, Shazeer, Parmar, Uszkoreit, Jones,
  Gomez, Kaiser and Polosukhin}]{28}
\bibinfo{author}{Vaswani, A.}, \bibinfo{author}{Shazeer, N.M.},
  \bibinfo{author}{Parmar, N.}, \bibinfo{author}{Uszkoreit, J.},
  \bibinfo{author}{Jones, L.}, \bibinfo{author}{Gomez, A.N.},
  \bibinfo{author}{Kaiser, L.}, \bibinfo{author}{Polosukhin, I.},
  \bibinfo{year}{2017}b.
\newblock \bibinfo{title}{Attention is all you need}.
\newblock \bibinfo{journal}{ArXiv} \bibinfo{volume}{abs/1706.03762}.
\bibitem[{Woo et~al.(2018)Woo, Park, Lee and Kweon}]{19}
\bibinfo{author}{Woo, S.}, \bibinfo{author}{Park, J.}, \bibinfo{author}{Lee,
  J.Y.}, \bibinfo{author}{Kweon, I.S.}, \bibinfo{year}{2018}.
\newblock \bibinfo{title}{Cbam: Convolutional block attention module}, in:
  \bibinfo{booktitle}{ECCV}.
\bibitem[{Xu et~al.(2015a)Xu, Ba, Kiros, Cho, Courville, Salakhutdinov, Zemel
  and Bengio}]{29}
\bibinfo{author}{Xu, K.}, \bibinfo{author}{Ba, J.}, \bibinfo{author}{Kiros,
  R.}, \bibinfo{author}{Cho, K.}, \bibinfo{author}{Courville, A.C.},
  \bibinfo{author}{Salakhutdinov, R.}, \bibinfo{author}{Zemel, R.},
  \bibinfo{author}{Bengio, Y.}, \bibinfo{year}{2015}a.
\newblock \bibinfo{title}{Show, attend and tell: Neural image caption
  generation with visual attention}, in: \bibinfo{booktitle}{ICML}.
\bibitem[{Xu et~al.(2015b)Xu, Ba, Kiros, Cho, Courville, Salakhutdinov, Zemel
  and Bengio}]{11}
\bibinfo{author}{Xu, K.}, \bibinfo{author}{Ba, J.}, \bibinfo{author}{Kiros,
  R.}, \bibinfo{author}{Cho, K.}, \bibinfo{author}{Courville, A.C.},
  \bibinfo{author}{Salakhutdinov, R.}, \bibinfo{author}{Zemel, R.},
  \bibinfo{author}{Bengio, Y.}, \bibinfo{year}{2015}b.
\newblock \bibinfo{title}{Show, attend and tell: Neural image caption
  generation with visual attention}, in: \bibinfo{booktitle}{ICML}.
\bibitem[{Yang et~al.(2016)Yang, Yang, Dyer, He, Smola and Hovy}]{30}
\bibinfo{author}{Yang, Z.}, \bibinfo{author}{Yang, D.}, \bibinfo{author}{Dyer,
  C.}, \bibinfo{author}{He, X.}, \bibinfo{author}{Smola, A.},
  \bibinfo{author}{Hovy, E.}, \bibinfo{year}{2016}.
\newblock \bibinfo{title}{Hierarchical attention networks for document
  classification}, in: \bibinfo{booktitle}{HLT-NAACL}.
\bibitem[{Yin et~al.(2016)Yin, Sch{\"u}tze, Xiang and Zhou}]{27}
\bibinfo{author}{Yin, W.}, \bibinfo{author}{Sch{\"u}tze, H.},
  \bibinfo{author}{Xiang, B.}, \bibinfo{author}{Zhou, B.},
  \bibinfo{year}{2016}.
\newblock \bibinfo{title}{Abcnn: Attention-based convolutional neural network
  for modeling sentence pairs}.
\newblock \bibinfo{journal}{Transactions of the Association for Computational
  Linguistics} \bibinfo{volume}{4}, \bibinfo{pages}{259--272}.
\bibitem[{Zhang et~al.(2017a)Zhang, Xu, Li, Zhang, Wang, Huang and
  Metaxas}]{38}
\bibinfo{author}{Zhang, H.}, \bibinfo{author}{Xu, T.}, \bibinfo{author}{Li,
  H.}, \bibinfo{author}{Zhang, S.}, \bibinfo{author}{Wang, X.},
  \bibinfo{author}{Huang, X.}, \bibinfo{author}{Metaxas, D.N.},
  \bibinfo{year}{2017}a.
\newblock \bibinfo{title}{Stackgan: Text to photo-realistic image synthesis
  with stacked generative adversarial networks}.
\newblock \bibinfo{journal}{2017 IEEE International Conference on Computer
  Vision (ICCV)} , \bibinfo{pages}{5908--5916}.
\bibitem[{Zhang et~al.(2017b)Zhang, Zhu, Lei, Shi, Wang and Li}]{22}
\bibinfo{author}{Zhang, S.}, \bibinfo{author}{Zhu, X.}, \bibinfo{author}{Lei,
  Z.}, \bibinfo{author}{Shi, H.}, \bibinfo{author}{Wang, X.},
  \bibinfo{author}{Li, S.}, \bibinfo{year}{2017}b.
\newblock \bibinfo{title}{S3fd: Single shot scale-invariant face detector}.
\newblock \bibinfo{journal}{2017 IEEE International Conference on Computer
  Vision (ICCV)} , \bibinfo{pages}{192--201}.
\bibitem[{Zhou et~al.(2019)Zhou, Liu, Liu, Luo and Wang}]{4}
\bibinfo{author}{Zhou, H.}, \bibinfo{author}{Liu, Y.}, \bibinfo{author}{Liu,
  Z.}, \bibinfo{author}{Luo, P.}, \bibinfo{author}{Wang, X.},
  \bibinfo{year}{2019}.
\newblock \bibinfo{title}{Talking face generation by adversarially disentangled
  audio-visual representation}.
\newblock \href{http://arxiv.org/abs/1807.07860}{\tt arXiv:1807.07860}.
\bibitem[{Zhu et~al.(2017)Zhu, Park, Isola and Efros}]{36}
\bibinfo{author}{Zhu, J.Y.}, \bibinfo{author}{Park, T.},
  \bibinfo{author}{Isola, P.}, \bibinfo{author}{Efros, A.A.},
  \bibinfo{year}{2017}.
\newblock \bibinfo{title}{Unpaired image-to-image translation using
  cycle-consistent adversarial networks}.
\newblock \bibinfo{journal}{2017 IEEE International Conference on Computer
  Vision (ICCV)} , \bibinfo{pages}{2242--2251}.

\end{thebibliography}



\end{document}